\documentclass{article}
\usepackage{ijcai16}

\usepackage{times}

\usepackage{epsfig}
\usepackage{graphicx}
\usepackage{amsmath}
\usepackage{amssymb}
\usepackage{color}
\usepackage{dsfont}
\usepackage{algorithm}
\usepackage[noend]{algpseudocode}
\usepackage{hhline}

\title{Neural Self Talk: Image Understanding via \\ Continuous Questioning and Answering }
\author{
 Yezhou Yang$^{1}$, Yi Li$^{2}$, Cornelia Fermuller$^{1}$, and Yiannis Aloimonos$^{1}$
\\
 $^{1}$University of Maryland, College Park, MD 20742\\
 $^{2}$Toyota Research Institute of North America, Ann Arbor, MI 48105
}
\begin{document}

\maketitle

\begin{abstract}
In this paper we consider the problem of continuously discovering image contents by actively asking image based questions and subsequently answering the questions being asked.
The key components include a Visual Question Generation (VQG) module and a Visual Question Answering module, in which Recurrent Neural Networks (RNN) and Convolutional Neural Network (CNN) are used.
Given a dataset that contains images, questions and their answers, both modules are trained at the same time, with the difference being VQG uses the images as input and the corresponding questions as output, while VQA uses images and questions as input and the corresponding answers as output.
We evaluate the self talk process subjectively using Amazon Mechanical Turk, which show effectiveness of the proposed method.

\end{abstract}
\section{Introduction} \label{sec:intro}

Acclaimed as ``one of the last cognitive tasks to be performed well by computers'' \cite{stork1998hal}, exploring and analyzing novel visual scenes is a journey of continuous discovery, which requires not just passively detecting objects and segmenting the images, but arguably more importantly, actively asking the right questions and subsequently closing the semantic loop by answering the questions being asked.

\begin{figure}[t]
\begin{center}
\includegraphics[width=0.83\columnwidth]{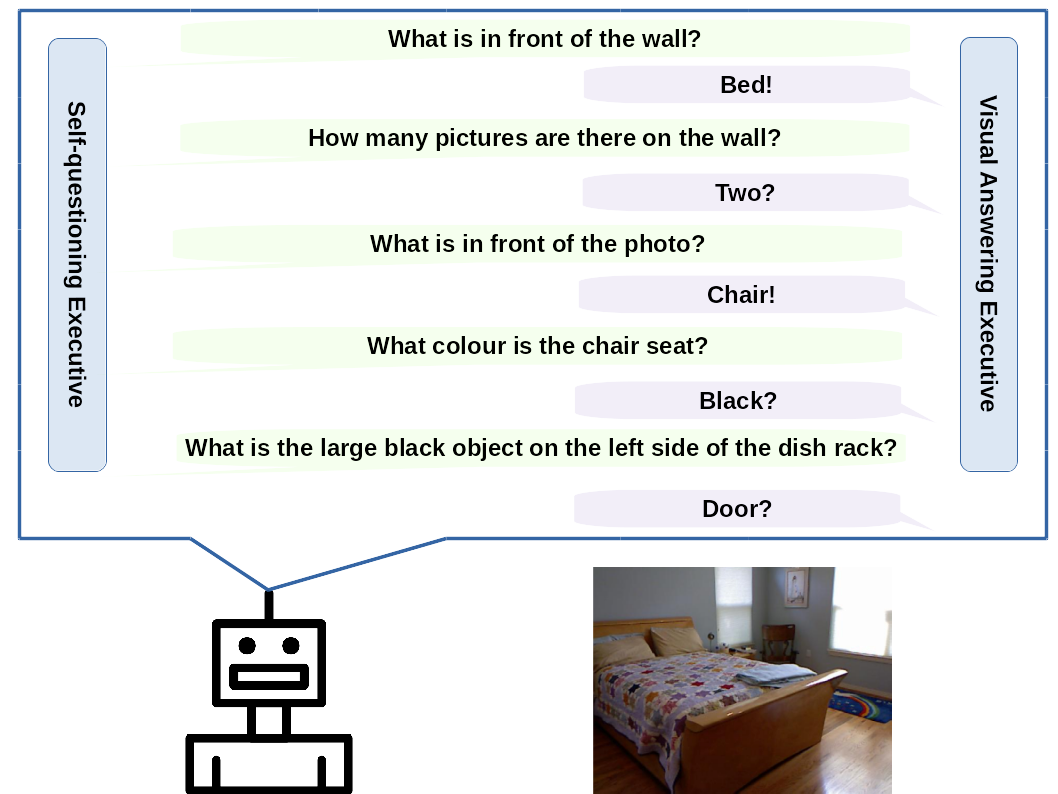}
\end{center}
\caption{One example self talk by the presented system, while the affirmative or questionable answer is decided by confidence score from visual answering executive.}
\label{fig:framework}
\end{figure}

This paper proposes a framework that can continuously discover novel questions on an image, and then provide legitimate answers.
This ``self talk'' approach for image understanding goes beyond visual classification by introducing a theoretically infinite interaction between a natural language question generation module and a visual question answering module. Under this architecture, the ``thought process'' for image understanding can be revealed by a sequence of consecutive question and answer pairs (Fig.~\ref{fig:framework}).  

Our ``self talk'' framework has two ``executives'' that takes their roles iteratively: 1) question generation, which is responsible for asking the right questions, and 2) question answering, which accepts the questions and generate potential answers.
With the rapid development in computer vision and machine learning \cite{mao2014explain,kiros2014unifying,donahue2014long,karpathy2014deep,vinyals2014show,chen2014learning} there are a few tools developed for this seemingly intuitive philosophy in Artificial Intelligence, but self-talk is certainly beyond the aggregation of tools, because it is fundamentally a challenging chicken egg problem.

\textbf{1) Questions from a single image can be as diversified as possible}.
Researchers have attempted a few approaches that mostly centered on asking limited questions such as ``what'' (\textit{e.g.,} object and action recognition) and ``where'' (\textit{e.g.,} place recognition).
Unfortunately, questions can be anything related or unrelated to the given picture.
This puzzling issue of unconstrained questions can be traced back to the original Turing test\footnote{
``Would  the  questions  have  to  be  sums,  or  could  I  ask  it what it had had for breakfast?'' Turing  ``Oh  yes,  anything.'' \cite{citeulike:335893}}, and the solution is still elusive.

Luckily, researchers have advanced the viewpoint that if we are able to develop a semantic understanding of a visual scene, we should be able to produce natural language descriptions of such semantics. This ``image captioning'' perspective are indeed exciting achievements, but it is only limited to generate descriptive captions, thus we propose to consider the question ``Can we generate questions, based on images?''.

\textbf{2) Evaluating the correctness of automatic questions answering is in the realm of Turing test}.
The ``Visual Question Answering'' \cite{VQA} problem recently becomes an important area in computer vision and machine learning, and sometimes it is referred as Visual Turing challenge \cite{malinowski2014towards}. A few approaches \cite{malinowski2015ask,ren2015image} have shown that deep neural nets again can be trained to answer a related question for an arbitrary scene with promising success.

\textbf{3) The semantic loop between the above two ``executives'' must be closed.} While the above two ``executives'' are very interesting entities, they cannot achieve the ``self talking'' by their own.
On one hand, the image captioning task neglects the importance of the thought process behind the appearance. Also, the amount of information covered by a finite language description is limited. These limitations have been pointed out by several recent works \cite{johnsoncvpr2015,schuster-EtAl:2015:VL,aditya2015images} and have been addressed partially by introducing middle layer knowledge representations. 
On the other hand, the setting of the visual question answering task requires as input a related question given by human beings. These questions themselves inevitably contain information about the image, which are recognized by human beings and only available through human intervention. Several recent results on the image VQA benchmarks indicate that language only information seem to contribute to the most of the good performance and how important the role of the visual recognition is still unclear.

In our formalism, the input of the final deep trained system is solely an image. Both questions and answers are generated from the trained models. Also, we want to argue that the capability of raising relevant and reasonable questions actively is the key to intelligent machinery.
Thus, the main contributions of this paper are twofold: 1) we propose to automatically generate ``self talk'' for arbitrary image understanding, a conceptually intuitive yet AI-challenging task; 2) we propose an image question generation module based on deep learning method.

\begin{figure}[t]
\begin{center}
\includegraphics[width=0.75\columnwidth]{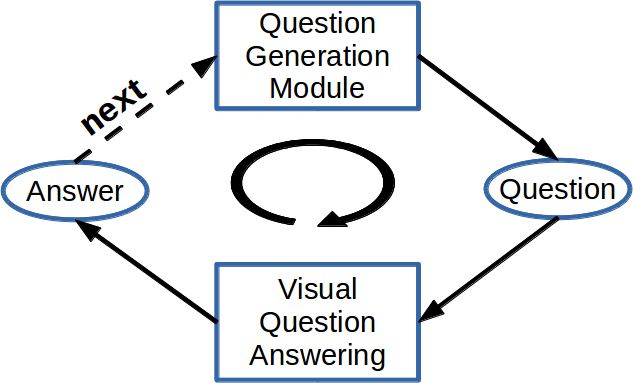}
\end{center}
\caption{The flow chart or our approach.}
\label{fig:arch}
\end{figure}

Figure.~\ref{fig:arch} illustrates the flow chart of our approach 
(Sec.\ref{sec:app}). 
In Sec.\ref{sec:exp}, we report experiments on two publicly available datasets (DAQUAR for indoor domain \cite{malinowski2014towards} and COCO for arbitrary domain \cite{VQA}). Specificaly, we 1) evaluate the quality of the generated questions using standard  language based metrics similar to image captioning and 2) use Amazon Mechanical Turk (AMT)-based evaluations of the generated question-answer pairs. We further discuss the insights from experimental results and challenges beyond them.

\section{Related Work}

Our work is related mainly to three lines of research of natural image understanding: 1) question generation, 2) image captioning and 3) visual question answering.

\textbf{Question Generation} is one of the key challenges in natural languages. Previous approaches of question generation from natural language sentences are mainly through template matching in a conservative manner \cite{brown2005automatic,heilman2010good,ali2010automation}. \cite{ren2015image} proposed to use parsing based approach to synthetically create question and answer pairs from image annotations. In this paper, we propose a visual question generation module through a technique directly adapted from image captioning system \cite{karpathy2014deep}, which is data driven and the potential output questions space is significantly larger than previous parsing or template based approaches, and the trained module only takes in image as input. 

In \textbf{Image Captioning}, in addition to the deep neural nets based approaches mentioned in Sec. \ref{sec:intro} we also share our roots with the works of generating textual descriptions. This includes the works that retrieves and ranks sentences from training sets given an image such as \cite{hodosh2013framing}, \cite{FarhadiPictureStory}, \cite{im2textOrdonez}, \cite{DBLP:journals/tacl/SocherKLMN14}. \cite{DBLP:conf/emnlp/ElliottK13}, \cite{Kulkarni11babytalk:}, \cite{Kuznetsova:2012:CGN:2390524.2390575}, \cite{Yang:2011:CSG:2145432.2145484}, \cite{journals/pieee/YaoYLLZ10} are some of the works that have generated descriptions by stitching together annotations or applying templates on detected image content.

In the filed of \textbf{Visual Question Answering}, very recently researchers spent a significant amount of efforts on both creating datasets and proposing new models \cite{VQA,malinowski2015ask,gao2015you,ma2015learning}. Interestingly both \cite{VQA} and \cite{gao2015you} adapted MS-COCO \cite{lin2014microsoft} images and created an open domain dataset with human generated questions and answers. The creation of these visual question answering testbed costs more than 20 people year of effort using Amazon Turk platform, and some questions are very challenging which actually require logical reasoning in order to answer correctly. Both \cite{malinowski2015ask} and \cite{gao2015you} use recurrent networks \cite{} to encode the sentence and output the answer. Specifically, \cite{malinowski2015ask} applies a single network to handle both encoding and decoding, while \cite{gao2015you} divides the task into an encoder network and a decoder one. More recently, the work from \cite{ren2015image} reported state-of-the-art VQA performance using multiple benchmarks. The progress is mainly due to formulating the task as a classification problem and focusing on the domain of questions that can be answered with one word. The visual question answering module adopts this approach.

\section{Self talk: Theory and Practice}\label{sec:app}

\subsection{Theory and Motivation}


The phenomenon of ``self talk'' has been studied in the field of psychology for hundreds of years. The term is defined as a special form of intrapersonal communication: a communicator's internal use of language or thought. Using the terms of computer science and engineering, it could be useful to envision intrapersonal communication occurring in the mind of the individual in a model which contains a sender, receiver, and a potential feedback loop. This process happens consciously or sub-consciously in our mind. 
The capability of self-raising questions and answer them is also crucial for learning.  Question raising and answering facilitate the learning process. For example, in the field of education, reciprocal questioning has been studied as a strategy, where students take on the role of the teacher by formulating their own list of questions about a reading material. 
In this paper, we regard this as another challenge for computers, and we believe that one key to intelligence is raising the right questions.

The benefits of modeling scene understanding task as a revealing of the ``self talk'' of the intelligent agents are mainly twofold: 1) the understanding of the scene can be revealed step by step and the failure cases could be tracked to specific question answer pairs. In other words, the process is more transparent; 2) theoretically the number of questions could be infinite and the question and answer loop could be never ending. This is especially crucial for active agent, such as movable robots, while their view of the scene keeps changing by moving around space, and the ``self talk'' in this scenario is never-ending. For a specific task, such as scene category recognition, this formulation has been proven to be efficient \cite{yu2011active}.

From a practical point of view, the revealing of ``self talk'' makes computers more human like, and the presented system has application potential in creating robotic companions \cite{yu2011active}. Note that as human being, we make mistakes, and some of them are ``cute'' mistakes. In Sec.~\ref{sec:exp}, we show that our system makes many ``cute'' mistakes too, which actually makes it more human-like. 

\subsection{Our Approach}


We have two hypotheses to validate in this work: 1) with the current progress in image captioning, a system can be trained to generate reasonable and relevant questions, and 2) by incorporating it with a visual question answering system, a system could be trained to generate human like ``self talk'' with promising success.

In this section, we introduce a frustratingly straightforward policy to generate a sequence of questions for the purpose of ``self talk''. We repeat this sampling process $q = QuestionSampling(I)$ $N$ times (five times typically in our experiments). For each question $q_i$ generated and the accompanied original image $I$, we pass it through the VQA module $a = VisualAnswer(q, I)$ to achieve an answer $a_i$. In such a manner we achieve the ``self talk'' question and answers pairs $\{(q_1,a_1),...,(q_N,a_N)\}$. The ``self talk'' is further evaluated by Amazon Mechanical Turk based human evaluation.

\begin{algorithm}
\caption{A Primitive ``Self Talk'' Generation Algorithm}\label{algo:inferConcepts}
{\fontsize{7}{7}\selectfont
\begin{algorithmic}[1]
\Procedure{SelfTalkGeneration}{($I$)}
\State $i \gets 1$
\While {$i \leq N$} 
\State $q_i = QuestionSampling(I)$
\State $a_i = VisualAnswer(q_i, I)$ 
\State $i = i + 1$
\EndWhile
\Return $\{(q_1,a_1),...,(q_N,a_N)\}$
\EndProcedure
\end{algorithmic}
}
\end{algorithm} 

\subsubsection{Question Generation}

In this section, we assume an input set of images and their questions raised by human annotators. In our scenario, these are full images and their questions set. We adopted the method from \cite{karpathy2014deep}, where a simple but effective extension is introduced from previously developed Recurrent Neural Networks (RNNs) \cite{} based language models to train image captioning model effectively. For the purpose of a self-contained work, we briefly go over the method here.

Specifically, during the training of our image question generation module, the multimodal RNN takes the image pixels $I$ and a sequence of input vectors $(x_1,...,x_T)$. It then computes a sequence of hidden states $(h_1,...,h_T)$ and a sequence of outputs $(y_1,...,y_T)$ by iterating the following recurrence relation from $t = 1$ to $t = T$. 
\begin{align}
 & b_v = W_{hi}[CNN_{\theta_c}(I)] \\
 & h_t = f(W_{hx} x_t + W_{hh} h_{t-1} + b_h + \mathds{1}(t = 1) \odot b_v ) \\
 & y_t = softmax(W_{ho}h_t + b_o),
\end{align}

In the equations above, $W_{hi}$, $W_{hx}$, $W_{hh}$, $W_{oh}$, $x_i$ and $b_h$,$b_o$ are learnable parameters, and $CNN_{\theta_c}(I)$ is the last layer of a pre-trained Convolutional Neural Network (CNN) \cite{}. The output vector $y_t$ holds the (unnormalized) log probabilities of words in the dictionary and one additional dimension for a special END token. In the approach, the image context vector $b_v$ to the RNN is only given at the first iteration. A typical size of the hidden layer of the RNN is 512 neurons.

The RNN is trained to combine a word $(x_t)$, the previous context $(h_{t−1})$ to predict the next word $(y_t)$ in the generated question. The RNN’s predictions on the image information $b_v$ via bias interactions on the first step. The training proceeds as follows (refer to Figure.\ref{fig:ques}a)): First set $h_0 = 0$, $x_1$ to a special START vector, and the desired label $y_1$ as the first word in the training question. Then set $x_2$ to the word vector of the first word and expect the network to predict the second word, etc. Finally, on the last step when $x_T$ represents the last word, the target label is set to a special END token. The cost function is to maximize the log probability assigned to the target labels (here, a Softmax classifier).

During testing time, to generate one question, we first compute the image representation $b_v$, and then set $h_0 = 0$, $x_1$ to the START vector and compute the distribution over the first word $y_1$ . We sample each word in the question from the distribution, set its embedding vector as $x_2$ , and repeat this process until the END token is generated. In the rest of the paper, we denote this question generation process as $q = QuestionSampling(I)$.

\subsubsection{Question Answering}

\begin{figure}[t]
\begin{center}
\includegraphics[width=\columnwidth]{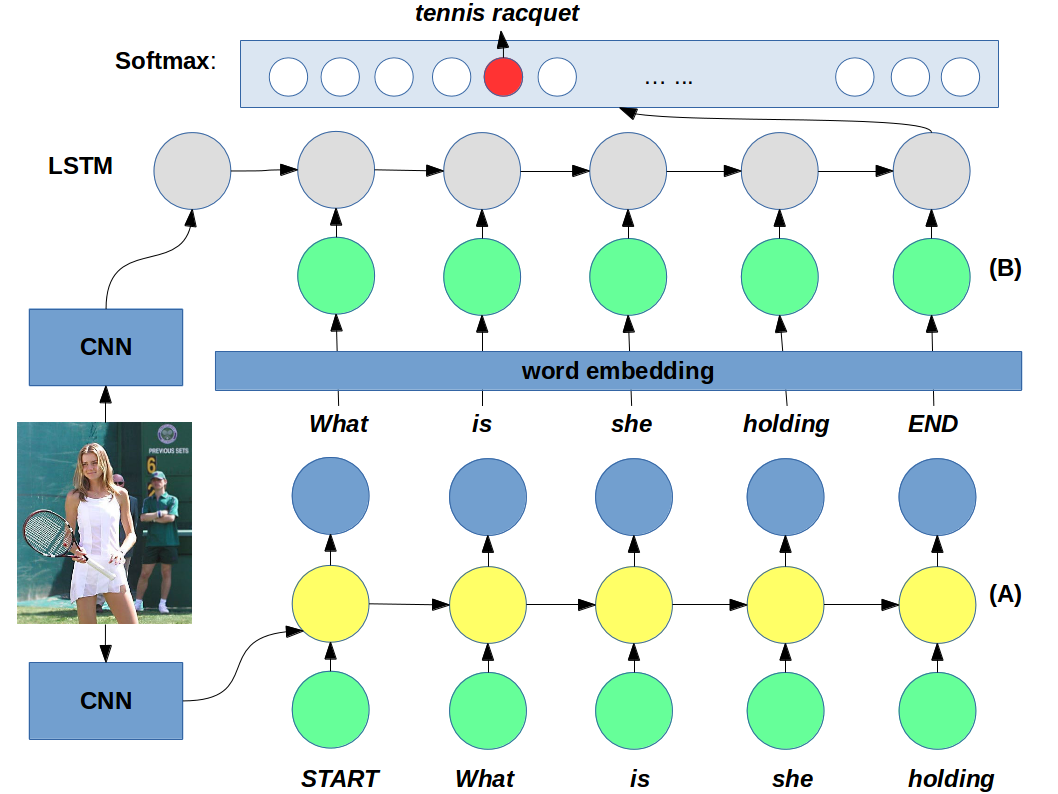}
\end{center}
\caption{The presented architecture of question generation module (part A) and question answering module (part B), and how they are connected.}
\label{fig:ques}
\end{figure}

In this section, we assume an input set of images and their annotated question answer pairs from human labelers. We adopted the approach from \cite{ren2015image}, which introduced a model builds directly on top of the long short-term memory (LSTM) \cite{hochreiter1997long} sentence model and is called the “VIS+LSTM” model. It treats the image as one word of the question as shown in Figure.\ref{fig:ques}b). 

The model uses the last hidden layer of the 19-layer Oxford VGG Conv Net \cite{simonyan2014very} trained on ImageNet 2014 Challenge as the visual embeddings. The CNN part of the model is kept frozen during training. The model also uses word embedding model from general purpose skip-gram embedding \cite{pennington2014glove}. In our experiments, the word embedding is kept dynamic (trained with the rest of the model).
Please refer to \cite{ren2015image} for the details.
In the rest of the paper, we denote this trained VQA module as $a = VisualAnswer(q, I)$.

\subsubsection{Amazon Mechanical Turk based Evaluation}
For the generated question answer (``self talk'') pairs, since there are no groundtruth annotations that could be used for automatic evaluation, we designed a Amazon Mechanical Turk (AMT) based human evaluation metric to report.

We ask the Turkers to imagine they have a companion robot whose name is ``self talker''. Once they bring the robot to a place shown in the image give, the robot started to generate questions and then self-answer the questions as if he is talking to himself. Specifically we ask the Turkers to evaluate three metrics: 1) Readability: how readable the ``self talker'' 's ``self talk''. Scores range from 1: not readable at all, to 5: no grammatical errors. Grammatically sound ``self-talk'' have better readability ratings; 2) Correctness: how correct the ``self talk'' is. ``self-talk'' content that correctly describes the image content with higher precision have better correctness ratings (range from 1 to 5); 3) Human likeness: how human-like does the robot perform (range from 1 to 5). 

\section{Experiments}
\label{sec:exp}

We test the presented approach on two visual question answering (VQA) datasets, namely, DARQUAR \cite{malinowski2014towards} and MSCOCO-VQA \cite{VQA}. In the experiments on these two datasets, we first report the question generation performance using standard image captioning language based evaluation metric. Then, in order to evaluate the performance of the ``self talk'' we report the AMT results and provide further discussion.


\subsection{Datasets}

We first briefly describe the two testing-beds we are using for the experiments.

\begin{figure*}[!ht]
\begin{center}
\includegraphics[width=\linewidth]{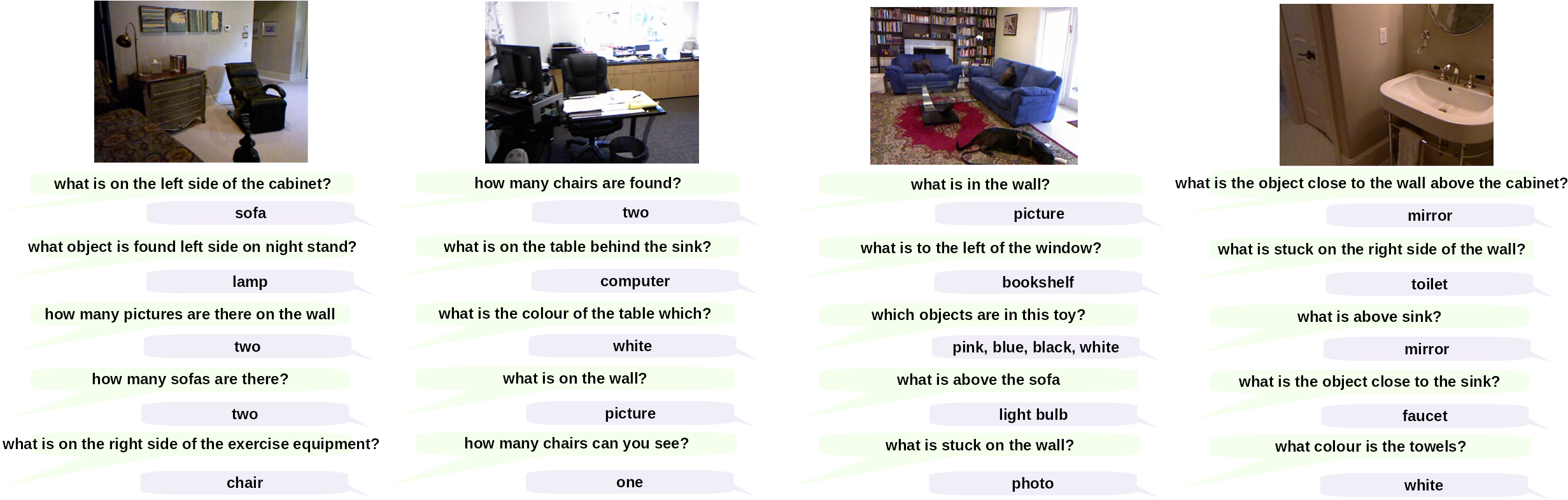}
\end{center}
\caption{Example ``self talk'' generated on DAQUAR testing set.}
\label{fig:exp2}
\end{figure*}

{\bf DAQUAR: Indoor Scenes}: DAQUAR \cite{malinowski2014towards} vqa dataset contains 12,468 human question answer pairs on 1,449 images of indoor scene. The training set contains 795 images and 6,793 question answer pairs, and the testing set contains 654 images and 5,675 question answer pairs. We run experiments for the full dataset with all classes, instead of their “reduced set” where the output space is restricted to only 37 object categories and 25 test images in total. This is because the full dataset is much more  challenging and the results are more meaningful in statistics. 

\begin{figure*}[!ht]
\begin{center}
\includegraphics[width=\linewidth]{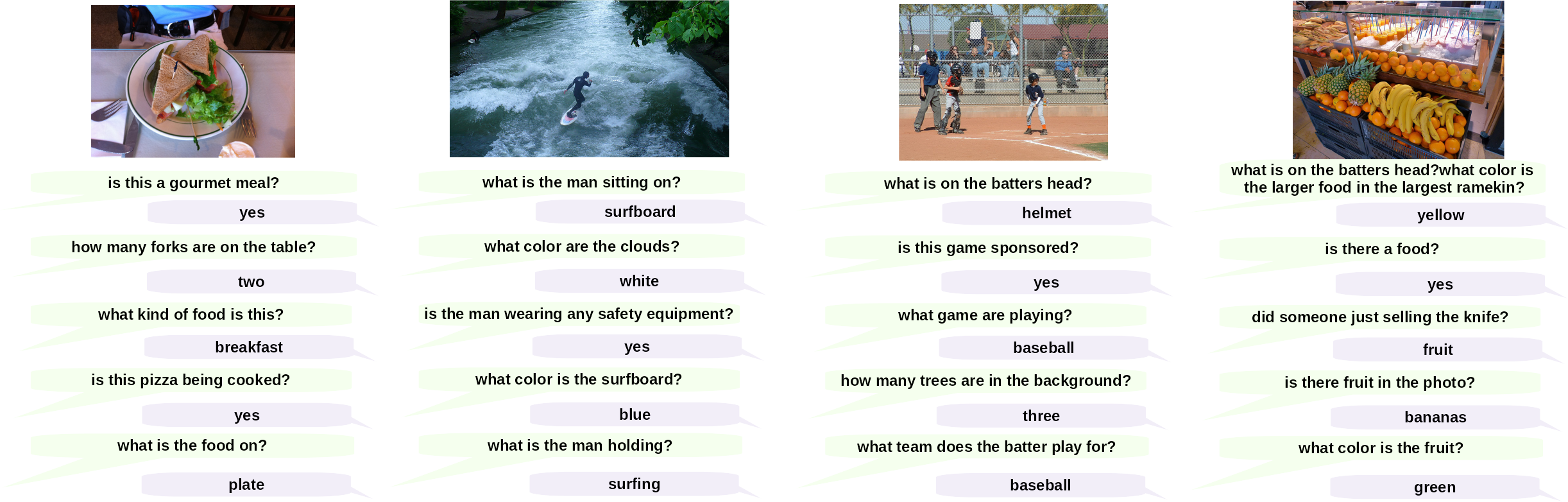}
\end{center}
\caption{Example ``self talk'' generated on MSCOCO-VQA testing set.}
\label{fig:exp1}
\end{figure*}

{\bf COCO: General Domain}: MSCOCO-VQA \cite{VQA} is the latest VQA dataset that contains open-ended questions about arbitrary images collect from the Internet. This dataset contains 369,861 questions and 3,698,610 ground truth answers based on 123,287 MSCOCO images. These questions and answers are sentence-based and open-ended. The training and testing split follows MSCOCO-VQA official split. Specifically, we use 82,783 images for training and 40,504 validation images for testing. The variation of the images in this dataset is large and till now it is considered as the largest general domian VQA dataset. The effort of collecting this dataset cost over 20 people year working time using Amazon Mechanical Turk interface. 

\subsection{Question Generation Evaluation}

We now evaluate the ability of our RNN model to raise questions about a given image. We first trained our Multimodal RNN to generate questions on full images with the goal of verifying that the model is rich enough to support the mapping from image data to sequences of words.  We report the BLEU \cite{papineni2002bleu}, METEOR \cite{lavie2014meteor}, ROUGE \cite{lin2004rouge} and CIDEr \cite{vedantam2014cider} scores computed with the coco-caption code \cite{chen2015microsoft}. Each method evaluates a candidate generated question by measuring how well it matches a set of several reference questions (averagely eight questions for DAQUAR dataset, and three questions for MSCOCO-VQA) written by humans.

\begin{table*}[!ht]
\centering

\begin{tabular}{|l|l|l|l|l|l|l|l|}
\hline
                & CIDEr & METEOR & ROUGE\_L & Bleu-1 & Bleu-2 & Bleu-3 & Bleu-4 \\ \hline
DAQUAR question MAX      & .512  & .361   & .761     & .81    & .735   & .635   & .361   \\ \hline
DAQUAR question SAMPLE   & .143  & .256   & .631     & .685   & .561   & .428   & .337   \\ \hline
coco-VQA question MAX    & .331 & .178  &  .493        &  .594      &    .422    & .291       & .193       \\ \hline
coco-VQA question SAMPLE & .133 & .127  &  .342 &  .388      & .220       &   .117     &   .064     \\ \hhline{|=|=|=|=|=|=|=|=|}
coco-Caption \cite{karpathy2014deep} & .66 & .195  & -- &  .625     & .45       &   .321     &   .23    \\ \hline
\end{tabular}
\caption{Evaluation of question generation on DAQUAR and coco-VQA datasets. MAX: the generated questions have max probability from trained model. SAMPLE: the generated questions are randomly drawn from the trained probabilistic model. }
\label{tb:caption}
\end{table*}

To further validate the performance of question generation, we further list the performance metrics reported in the state-of-the-art image captioning work \cite{karpathy2014deep}. From Table.~\ref{tb:caption}, except CIDEr score, the question generation performance is comparable with the state-of-the-art image captioning performance. Note that for CIDEr score is a consensus based metric. The facts that, 1) coco-VQA has three reference ground-truth questions while coco-Caption has five and 2) human annotated questions by its nature varies more than captions, makes it hard to achieve high CIDEr score for question generation task.   

\subsection{``Self talk'' Evaluation}

\begin{figure*}[!htbp]
\begin{center}
\includegraphics[width=0.95\linewidth]{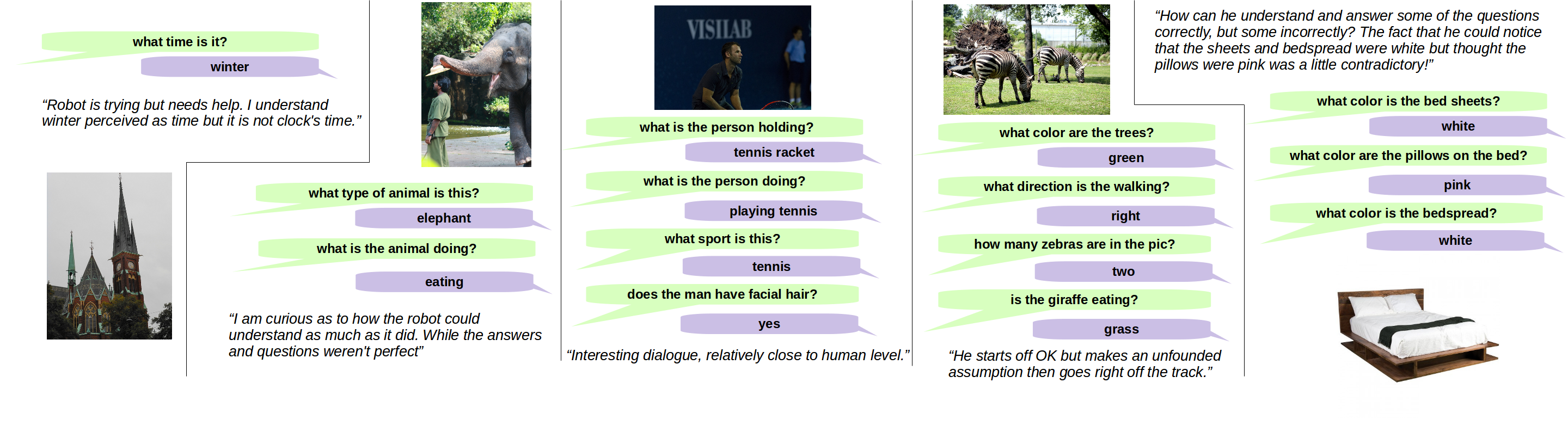}
\end{center}
\caption{Example Turkers' comments about the ``self talk'' robot.}
\label{fig:feedback}
\end{figure*}

In Table.~\ref{tb:amt} we report the average score as well as its standard deviation for each metric. We randomly drawn 100 and 1000 testing samples from DAQUAR and MSCOCO-VQA testing sets for the human evaluation reported here. From the human evaluation, we can see that the questions generated have achieved close to human readability. The correctness of the generated ``self talk'' averagely has some relevance to the image and according to Turkers, the imagined companion robot acts averagely beyond ``a bit like human being'' but below the ``half human, half machine'' category. 

\begin{figure}[!htbp]
\begin{center}
\includegraphics[width=\linewidth]{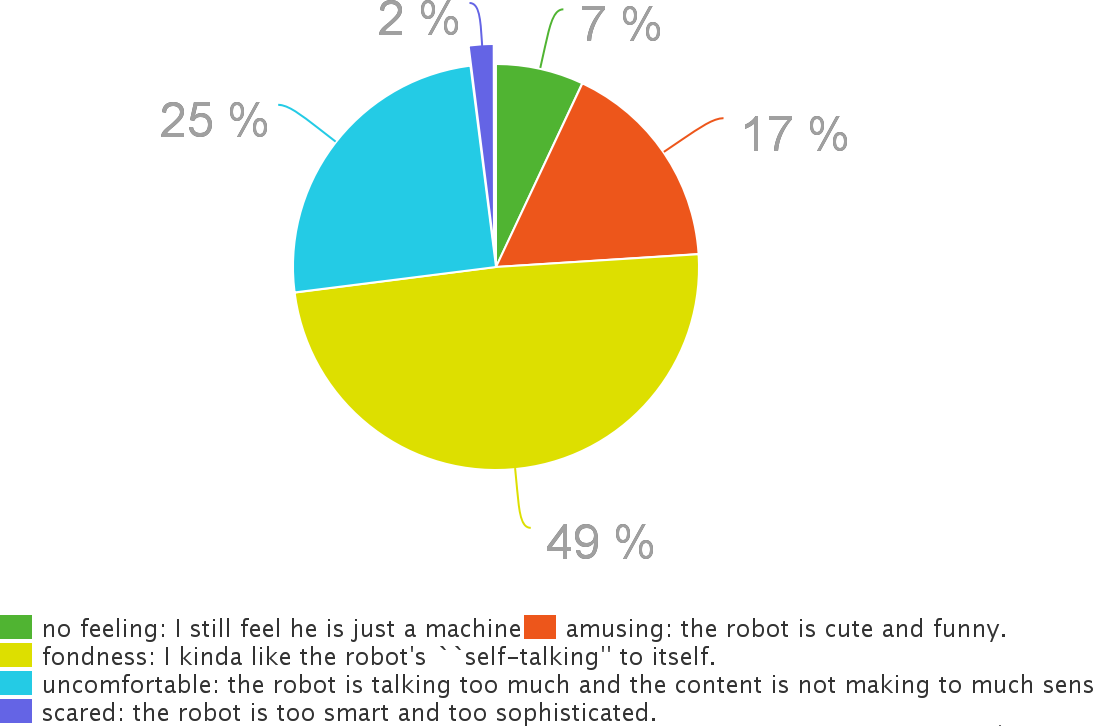}
\end{center}
\caption{User feedback on DAQUAR set.}
\label{fig:daq_amt}
\end{figure}

\begin{figure}[!htbp]
\begin{center}
\includegraphics[width=\linewidth]{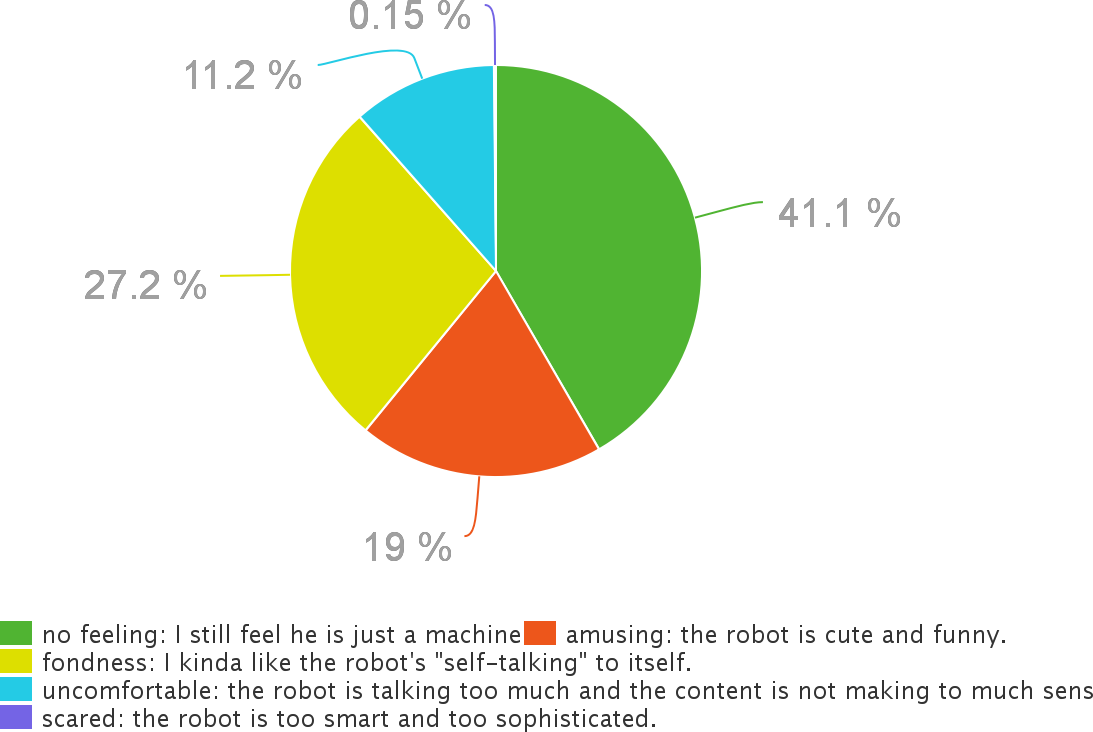}
\end{center}
\caption{User feedback on MSCOCO-VQA set.}
\label{fig:coco_amt}
\end{figure}

We also asked the Turkers to choose from five immediate feelings after their companion robot's performance. Fig.~\ref{fig:daq_amt} and Fig.~\ref{fig:coco_amt} depicts the feedback we got from the users. Given the fact that the performance of the ``self talker'' robot is still far from human performance, most of the Turkers thought they like such a robot or feel its amusing. And only very few of the users felt scared, which indicates that our image understanding performance is far from being trapped into the so-called ``uncanny valley'' \cite{mori2012uncanny} of machine intelligence. At the end of our evaluation, we also asked Turkers to comment about what the robot's performance. Some example comments could be found in Fig.~\ref{fig:feedback}.

\begin{table}[!htbp]
\centering

\begin{tabular}{|l|l|l|l|}
\hline
         & Readability & Correctness & \begin{tabular}[c]{@{}l@{}}Human \\ likeness\end{tabular} \\ \hline
DAQUAR   & $3.35\pm0.92$          & $2.5\pm1.03$           & $2.49\pm1.02$                                                         \\ \hline
coco-VQA & $3.39\pm1.18$          & $2.7\pm1.29$           & $2.40\pm1.33$                                                         \\ \hline
\end{tabular}
\caption{``Self talk'' AMT human evaluation.}
\label{tb:amt}
\end{table}

\section{Conclusion and Future Work}
\label{sec:con}

In this paper, we consider the image understanding problem as a self-questioning and answer process and we present a primitive ``self talk'' generation method based on two deep neural network modules. From the experimental evaluation on both the performance of question generation and final ``self talk'' pairs, we show that the presented method achieved a decent amount of success. There are still several potential pathways to improve the performance of intelligent ``self talk''.

{\bf The role of common-sense knowledge.} Common-sense knowledge has a crucial role in question raising and answering process for human beings \cite{aditya2015images}. The experimental result shows that our system by learning the model from large annotated question answer pairs, it implicitly encodes a certain level of common-sense. The real challenge is to deal with situations that the visual input conflicts with the common-sense learned from context data. In our experiment, it seems that the model is biased towards to trust his common sense more than the visual input. How to incorporate either logical or numerical forms of common-sense into end-to-end based image understanding system is still an open problem. 

{\bf Creating a story-line. } When human beings perform  intrapersonal communication, we tend to follow a logic flow or so-called story-line. This requires a question generation modules that takes in consideration the answers from previous questions for consideration. This indicates a more sophisticated dialogue generation process (such as a cognitive dialogue \cite{aloimonos2015cognitive}), and it can also potentially prevent self-contradictions happened in this paper's generated results (see last comment in Fig.~\ref{fig:feedback}).

As indicated from AMT feedback, human users felt it is cute and fondness to have a robot companion that moves around and talkative. Another open avenue is to integrate the current trained model onto a robot platform and through interaction with users to continuously refine its trained model.



\clearpage
\bibliographystyle{named}
\bibliography{vision_nlp}

\end{document}